\documentclass{article}

\usepackage[T1]{fontenc}
\usepackage[preprint]{neurips_2026}

\usepackage{graphicx}
\usepackage{booktabs}
\usepackage{amsmath}
\usepackage{amssymb}
\usepackage{xcolor}
\usepackage{microtype}
\usepackage{caption}

\newcommand{\nEpisodes}{2{,}000}
\newcommand{\ceilCal}{0.992}
\newcommand{\ceilPhys}{0.988}
\newcommand{\ceilAction}{0.524}
\newcommand{\NzceilCal}{0.992}

\newcommand{\NzceilAction}{0.514}
\newcommand{\auditDead}{13}
\newcommand{\auditTotal}{20}

\newcommand{\DinovProbeF}{0.291}
\newcommand{\DinovNSeventyFive}{0.720}

\newcommand{\FramediffProbeM}{0.986}

\newcommand{\FramediffNSeventyFive}{0.833}

\newcommand{\FramediffEraseM}{-0.306}

\newcommand{\PixelCal}{0.996}

\newcommand{\PixelProbeM}{0.983}

\newcommand{\PixelEraseM}{-0.001}

\newcommand{\RthreemNSeventyFive}{0.862}

\newcommand{\RthreemEraseM}{-0.004}

\newcommand{\RandvitCal}{0.977}

\newcommand{\RandvitProbeF}{0.361}
\newcommand{\RandvitNSeventyFive}{0.741}

\newcommand{\VconeProbeF}{0.390}
\newcommand{\VconeNSeventyFive}{0.765}

\newcommand{\VideomaeNSeventyFive}{0.931}

\newcommand{\VjepaCal}{0.995}

\newcommand{\VjepaNSeventyFive}{0.876}

\newcommand{\NzDinovCal}{0.673}

\newcommand{\NzDinovNSeventyFive}{0.167}

\newcommand{\NzFramediffCal}{0.944}

\newcommand{\NzFramediffNSeventyFive}{0.595}

\newcommand{\NzPixelCal}{0.861}

\newcommand{\NzPixelNSeventyFive}{0.334}

\newcommand{\NzPixelEraseM}{-0.350}

\newcommand{\NzRthreemCal}{0.703}

\newcommand{\NzRthreemNSeventyFive}{0.143}

\newcommand{\NzRthreemEraseM}{-0.208}

\newcommand{\NzRandvitCal}{0.476}

\newcommand{\NzRandvitNSeventyFive}{0.017}

\newcommand{\NzVconeCal}{0.712}

\newcommand{\NzVconeNSeventyFive}{0.176}

\newcommand{\NzVideomaeCal}{0.912}

\newcommand{\NzVideomaeNSeventyFive}{0.405}

\newcommand{\NzVjepaCal}{0.974}

\newcommand{\NzVjepaNSeventyFive}{0.716}

\newcommand{\SeqRandvitProbeF}{0.672}

\newcommand{\PlanOracleErr}{2.9}

\newcommand{\PlanTol}{10}

\newcommand{\NzPlanRandvitErr}{19.6}

\newcommand{\NzPlanVjepaErr}{4.0}
\newcommand{\NzPlanVjepaHit}{82}

\newcommand{\NzPlanBlindErr}{19.0}

\newcommand{\spreadClean}{0.020}
\newcommand{\spreadCleanTwo}{0.02}
\newcommand{\spreadNz}{0.498}
\newcommand{\spreadNzTwo}{0.50}
\newcommand{\maxRandomCtl}{0.004}
\newcommand{\NzmaxRandomCtl}{0.006}

\newcommand{\AggRandvitCal}{+0.000}

\newcommand{\dead}{\textcolor{gray!85}{\textbf{--}}}

\title{CALIPER: Clean Scenes Cannot Rank Physical Inference in Pretrained Visual Representations}

\author{%
  Aman Mehta\thanks{\texttt{heyamanmehta@gmail.com}} \\
  Independent Researcher
  \And
  Riya Baviskar\thanks{\texttt{riyakbaviskar@gmail.com}} \\
  Independent Researcher
}

\begin{document}
\maketitle


\begin{abstract}
How far a pushed object slides depends on its mass and friction, which no single image reveals. Pretrained visual encoders are the usual perception front end of world models for manipulation, and their grasp of such latent physics is judged with perturbation benchmarks and linear probes, almost always in a clean, fixed-camera scene. We show that these evaluations cannot tell an encoder that infers physics from one that does not. CALIPER (calibrate, then predict) strikes an object of unknown mass and friction twice at known speeds, shows a third strike only up to the moment of contact, and asks a linear readout on frozen features to predict the slide; swapping in another object's calibration clips checks that the evidence is used. Across \nEpisodes{} simulated episodes and eight representations, calibration adds $+0.50$ $R^2$ and the swap removes it, yet in the clean scene every representation, a randomly initialised ViT included, lands within \spreadCleanTwo{} $R^2$ of the simulator-state ceiling, because a fixed camera exposes the displacement directly in pixel coordinates. Resampling camera, lighting, and clutter for every clip spreads the same representations across \spreadNzTwo{} $R^2$; asked to choose a push speed for a goal distance, V-JEPA~2 misses by \NzPlanVjepaErr{}\,mm and the random ViT by \NzPlanRandvitErr{}\,mm, no better than ignoring the object. Linear probes track none of this: a change in frame aggregation moves a probe more than pretraining does, and erasing the probed mass direction from the same representation costs nothing in one scene and $0.35$ $R^2$ in the other. A benchmark's ability to rank models has to be measured, and we give three checks that do so.
\end{abstract}

\section{Introduction}

A light box and a heavy one of the same shape look identical until something moves them. A world model that is meant to support manipulation therefore has to infer mass and friction from the interactions it has watched, and then use that inference to predict what an action it has not yet seen will do. Pretrained visual encoders such as V-JEPA~2, VC-1, and R3M have become the standard front end for such models, and they are chosen on the strength of evaluations that are supposed to measure this ability. The two most common are single-parameter perturbation, where a scene is rendered twice with one property changed and a large change in the representation is read as physical sensitivity, and probing, where a linear map is fit from activations to the true property. Both are almost always run in a scene with one object, a fixed camera, and nothing else in view.

We find that neither test, in that scene, can tell an encoder that infers physics from one that does not. They give nearly the same score to every encoder we tried, including one whose weights were never trained. The encoders are not at fault; the evaluations were simply never checked for whether they could rank anything.

\begin{figure}[t]
\centering
\includegraphics[width=0.8\linewidth]{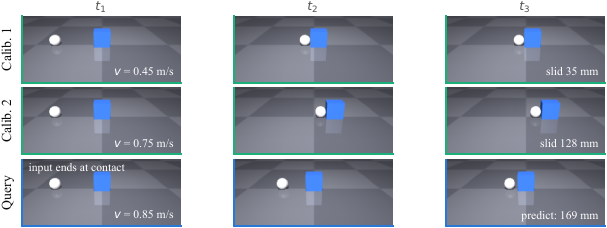}
\caption{One CALIPER episode. A standard puck strikes a target of unknown mass and friction at two known speeds (calibration), and the resulting slides reveal the target's physics. A third collision at a new speed is the query; the model's input stops at the moment of contact, so it contains the action but not its consequence. The model predicts the displacement that follows. Appearance is sampled independently of mass and friction.}
\label{fig:protocol}
\end{figure}

Our test, CALIPER (\emph{calibrate, then predict}), is built so that the answer is not in the input (Figure~\ref{fig:protocol}). A target of unknown mass and friction is struck twice by a standard puck at known speeds, and the two resulting slides, the \emph{calibration} clips, are shown in full. A third strike at a new speed, the \emph{query}, is shown only up to the moment of contact, and the model has to predict how far the object will slide. Appearance is sampled independently of physics, so no frame gives the answer away. Because the calibration clips are separate inputs we can also \emph{swap} them for another object's; any accuracy that survives the swap did not come from physical inference. We run the task in two scenes, one with a fixed camera and an empty table and one in which camera pose, lighting, floor colour, and two distractor objects are resampled for every clip, and we compare eight representations under one linear readout: V-JEPA~2, VideoMAE, VC-1, DINOv2, R3M, a randomly initialised ViT, raw pixels, and frame differences. Every claim is checked against simulator state that no learned model receives.

The test itself behaves as intended. Calibration raises $R^2$ by $+0.50$ for every representation and the swap removes all of it, so the readout is using the object's own interaction history. What the clean scene then shows is that this is not enough to rank anything: the eight representations span only \spreadClean{} $R^2$, raw pixels equal V-JEPA~2, and the random ViT is within \spreadClean{} of both. The reason is mechanical. With a fixed camera the object's displacement becomes a fixed pattern of pixel coordinates, and any injective feature map preserves that pattern, so a linear readout recovers the slide from almost anything. A perturbation benchmark we audit turns out to have the opposite problem: in \auditDead{} of \auditTotal{} of its task-property cells the perturbation moves the simulated outcome by less than $10^{-4}$, so there was nothing to measure. Once camera, lighting, and clutter vary from clip to clip, the same task does separate the representations, and by a wide margin: the spread grows to \spreadNz{} $R^2$, V-JEPA~2 keeps \NzVjepaCal{}, the per-frame encoders drop to about $0.7$, and the random ViT loses its calibration gain entirely. Asked to choose a push speed that will reach a goal distance, V-JEPA~2 misses by \NzPlanVjepaErr{}\,mm and the random ViT by \NzPlanRandvitErr{}\,mm, which is what you get by ignoring the object. Linear probes track none of this. Concatenating frames instead of averaging them raises the random ViT's friction probe from \RandvitProbeF{} to \SeqRandvitProbeF{}, above VC-1, without changing its prediction at all, and erasing the probed mass direction from raw pixels costs $\PixelEraseM{}$ $R^2$ in the clean scene but $\NzPixelEraseM{}$ under nuisance variation, from the same probe accuracy.

Our conclusion is that a benchmark's ability to separate representations has to be demonstrated rather than assumed. We end with three checks that demonstrate it, none of which the evaluations we started from would have passed.

\section{Related Work}

\textbf{How visual encoders for robotics are judged.} VC-1~\citep{vc1}, R3M~\citep{r3m}, V-JEPA~2~\citep{vjepa2}, VideoMAE~\citep{videomae}, and DINOv2~\citep{dinov2} are compared by downstream control success, which mixes perception, physical inference, and policy optimisation into one number. Freezing the encoder and fixing the readout isolates the middle term, at the cost of measuring only what is linearly available.

\textbf{Physics benchmarks.} IntPhys~\citep{intphys}, Physion~\citep{physion}, and PHYRE~\citep{phyre} score whether a model's physical predictions are correct. None reports whether its design could have ranked the models it compares: what a trivial baseline scores, or whether the perturbations it applies change the outcome. Those are the questions we ask.

\textbf{Identifying physics from motion.} Inferring object properties from how they move is a classical problem~\citep{galileo, physics101}, and the same idea underlies learning by poking~\citep{agrawal2016poke} and rapid adaptation from a short proprioceptive history~\citep{kumar2021rma}. We turn it into an evaluation. The calibration clips are the identification opportunity, and the swap tests whether it was taken.

\textbf{Decodability versus use.} Control tasks and amnesic probing established, for language models, that a property being linearly decodable from a representation does not mean the model uses it~\citep{hewitt2019control, belinkov2022probing, elazar2021amnesic}, and stated reasoning can diverge from actual reasoning~\citep{turpin2024language}. We bring the erasure test to visual physics and add one observation: the erasure cost of the same property from the same representation changes with the scene, so decodability is uninformative about use even within a single model.

\section{Calibration-Conditioned Prediction}
\label{sec:protocol}

\textbf{Environment.} Each episode contains one target: a box with mass $m \in [0.15, 0.75]$\,kg (log-uniform), sliding friction $\mu \in [0.12, 0.32]$, and half-extents, colour, and aspect ratio drawn independently of $(m, \mu)$, simulated in MuJoCo~\citep{mujoco}. A puck of fixed mass strikes the target head-on at a commanded speed. We use collisions rather than applied forces because a fixed force leaves objects below the friction threshold stationary and sends the rest out of frame, whereas collision recoil is bounded by the impact speed for any mass ratio (Appendix~\ref{app:design}). Two calibration collisions at $0.45$ and $0.75$\,m/s are shown in full. The query collision, at a speed drawn from $[0.40, 0.85]$\,m/s, is truncated at first contact. The regression target is $\log$ displacement during the query.

\textbf{Two scenes.} In the \emph{clean} scene the camera, lighting, and floor are fixed and the table holds only the puck and the target. In the \emph{nuisance} scene camera pose, lighting, floor tint, and two distractor objects are resampled for every clip, so the three clips of an episode do not share a frame of reference; each draw is checked to keep the workspace in view. Physics, the distribution of $(m, \mu)$, and the readout are identical across the two scenes. Anything that changes between them is a property of the scene, not of the task.

\textbf{Readout.} A frozen encoder maps each clip to one vector. Each clip's vector is centred and reduced by PCA on the training rows, standardised, concatenated with the commanded query speed, and fit by ridge regression with the penalty chosen by inner cross-validation, under repeated 5-fold cross-validation. All conditions share one component budget, split equally across clips, so adding a clip never adds parameters. Four conditions nest: \emph{action only} (the commanded speed); \emph{query} (adds the truncated query clip); \emph{query$+$calibration} (adds the object's own calibration clips); and \emph{query$+$swapped calibration} (another episode's clips, assigned by a derangement). We report $R^2$ on held-out episodes and do not interpret differences of a few hundredths, since repeated $K$-fold understates variance. Simulator state sets the reference points: the true calibration displacements reach $R^2=\ceilCal{}$, the true $(m,\mu)$ reach \ceilPhys{}, and the commanded speed alone reaches \ceilAction{}. The gap between the last and the first is the headroom that physical inference can fill. On the same folds we fit linear probes for $\log m$ and $\mu$ from the calibration features, and we remove a decoded property by iterative nullspace projection~\citep{ravfogel2020null, elazar2021amnesic} to test whether prediction depends on it.

\section{Results}
\label{sec:results}

\begin{table}[t]
\centering
\caption{Query-outcome $R^2$ on \nEpisodes{} episodes. \emph{Query}: truncated query clip only; \emph{$+$cal.}: adds the object's own calibration clips; \emph{$+$swap}: adds another object's. The commanded speed alone gives \ceilAction{} (clean) and \NzceilAction{} (nuisance); true simulator state gives \ceilCal{} and \NzceilCal{}. \emph{Goal err.}: median miss when the readout selects the push speed for a goal distance and the simulator executes it (Table~\ref{tab:plan}). The last row is the spread of the $+$cal.\ column, which is what a benchmark needs before it can rank anything.}
\label{tab:main}
\small
\begin{tabular}{lccc ccc cc}
\toprule
& \multicolumn{3}{c}{Fixed camera, clean scene} & \multicolumn{3}{c}{Varied camera, lighting, clutter} & \multicolumn{2}{c}{Goal err.\ (mm)} \\
\cmidrule(lr){2-4} \cmidrule(lr){5-7} \cmidrule(lr){8-9}
Representation & Query & +\,cal. & +\,swap & Query & +\,cal. & +\,swap & clean & nuis. \\
\midrule
V-JEPA 2 & 0.495 & 0.995 & 0.503 & 0.472 & 0.974 & 0.470 & 2.8 & 4.0 \\
VideoMAE & 0.492 & 0.994 & 0.501 & 0.480 & 0.912 & 0.474 & 2.8 & 7.1 \\
VC-1 & 0.486 & 0.985 & 0.495 & 0.476 & 0.712 & 0.470 & 3.2 & 13.2 \\
DINOv2 & 0.487 & 0.980 & 0.484 & 0.466 & 0.673 & 0.476 & 4.0 & 13.5 \\
R3M & 0.489 & 0.990 & 0.492 & 0.479 & 0.703 & 0.472 & 2.8 & 12.7 \\
Random ViT & 0.483 & 0.977 & 0.491 & 0.484 & 0.476 & 0.476 & 4.1 & 19.6 \\
Frame diff. & 0.500 & 0.996 & 0.500 & 0.476 & 0.944 & 0.473 & 2.7 & 5.7 \\
Raw pixels & 0.496 & 0.996 & 0.496 & 0.480 & 0.861 & 0.477 & 2.7 & 8.5 \\
\emph{True $(m,\mu)$} &  &  &  &  &  &  & 2.9 & 2.8 \\
\emph{Ignore object} &  &  &  &  &  &  & 18.0 & 19.0 \\
\midrule
\emph{Range of +\,cal.} &  & 0.020 &  &  & 0.498 &  &  &  \\
\bottomrule
\end{tabular}
\end{table}

\textbf{The clean scene.} Every representation gains about $+0.50$ $R^2$ from the object's own calibration clips and loses all of it when those clips are swapped for another object's (Table~\ref{tab:main}, left). The query clip alone adds nothing over the commanded speed, which confirms that appearance carries no usable physics. The swap is what lets us say the evidence is \emph{used}: the readout is not keying on anything that survives replacing it. What the clean scene cannot do is separate the representations. The with-calibration column spans only \spreadClean{}, with raw pixels at \PixelCal{}, V-JEPA~2 at \VjepaCal{}, and a ViT with random weights at \RandvitCal{}. We attribute this to the scene rather than to the encoders. With a fixed camera, the target's position after each calibration slide is a fixed function of pixel coordinates, so a linear readout can recover the slide distance from raw pixels directly, and equally from any injective transform of them, a random projection included; nothing has to be inferred. The perturbation benchmark audited in Appendix~\ref{sec:audit} fails in the opposite way. Simulated with no model at all, \auditDead{} of \auditTotal{} of its cells move the outcome by less than $10^{-4}$, so a sensitivity score computed there is fitting integrator noise. Both defects could have been seen before any model was run.

\begin{figure}[t]
\centering
\includegraphics[width=0.85\linewidth]{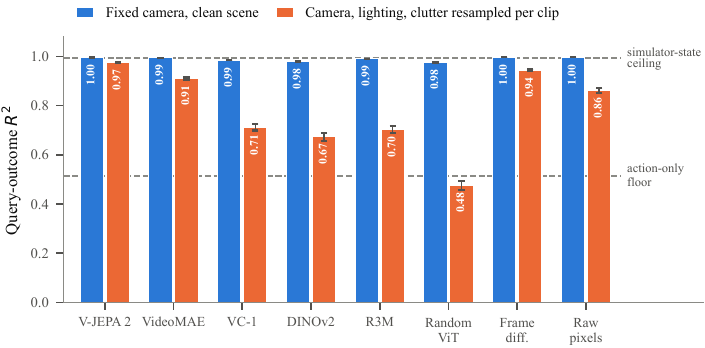}
\caption{Same task, two scenes. With a fixed camera and clean table (blue) every representation reaches the ceiling. With camera, lighting, floor, and distractors resampled per clip (orange) the same representations spread across \spreadNz{} $R^2$, and the random ViT drops to the floor set by the commanded speed alone.}
\label{fig:nuisance}
\end{figure}

\begin{figure}[t]
\centering
\includegraphics[width=\linewidth]{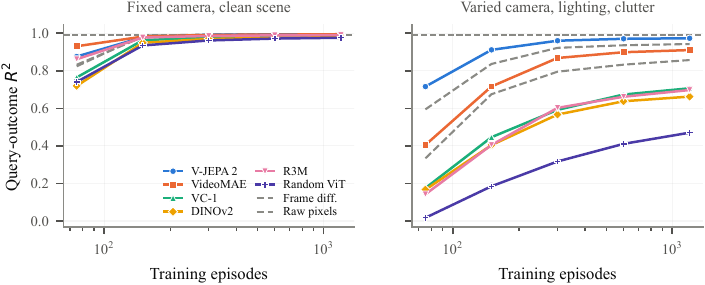}
\caption{Query-outcome $R^2$ against training-set size. Left: in the clean scene every representation converges to the same plateau. Right: under nuisance variation the ordering of Table~\ref{tab:main} holds at every training size.}
\label{fig:curve}
\end{figure}

\textbf{The nuisance scene.} Resampling camera, lighting, floor, and distractors for every clip changes neither the physics, the labels, nor the readout, yet the spread of the with-calibration column grows from \spreadClean{} to \spreadNz{} (Table~\ref{tab:main}, right; Figure~\ref{fig:nuisance}). The swap still removes every gain, so the ordering that appears reflects physical inference rather than some new shortcut. V-JEPA~2 retains \NzVjepaCal{}, frame differences \NzFramediffCal{}, VideoMAE \NzVideomaeCal{}, and raw pixels \NzPixelCal{}; VC-1, R3M, and DINOv2 fall to \NzVconeCal{}, \NzRthreemCal{}, and \NzDinovCal{}; the random ViT reaches \NzRandvitCal{} against a speed-only floor of \NzceilAction{}, so its features no longer separate an object's displacement from a camera move. The ordering holds at every training-set size (Figure~\ref{fig:curve}), and it groups by pretraining objective: the two models trained to predict video lead, the pixel-space references, which encode motion but no invariance, come next, and the three per-frame encoders trail regardless of what they were trained on. R3M and VC-1 are trained on the same egocentric video with different objectives and land within $0.01$ of each other here, which we take as weak evidence that data alone does not confer the needed invariance. One result surprised us: frame differences (\NzFramediffCal{}) beat VideoMAE (\NzVideomaeCal{}) and every per-frame encoder. We suspect the per-clip camera change is small enough that a $32\times32$ difference image still localises the moving object, and we have not tested larger viewpoint changes. In other words, what the clean scene reported as ``all representations are equivalent'' was a statement about the scene, not about the representations.

\textbf{Data efficiency.} In the clean scene only the approach to the plateau differs (Figure~\ref{fig:curve}, left). At 75 training episodes VideoMAE reaches \VideomaeNSeventyFive{}, V-JEPA~2 \VjepaNSeventyFive{}, R3M \RthreemNSeventyFive{}, frame differences \FramediffNSeventyFive{}, VC-1 \VconeNSeventyFive{}, DINOv2 \DinovNSeventyFive{}, and the random ViT \RandvitNSeventyFive{}. The R3M and VC-1 pair, matched on data and differing in objective, separates by $0.10$ here and by $0.40$ in friction decodability. We do not build a claim on this: these are differences between 75-sample fits whose fold spreads understate variance, and under nuisance variation the same pair is within $0.01$ at every size. The nuisance ordering at 75 episodes (V-JEPA~2 \NzVjepaNSeventyFive{}, frame differences \NzFramediffNSeventyFive{}, VideoMAE \NzVideomaeNSeventyFive{}, raw pixels \NzPixelNSeventyFive{}, VC-1 \NzVconeNSeventyFive{}, DINOv2 \NzDinovNSeventyFive{}, R3M \NzRthreemNSeventyFive{}, random ViT \NzRandvitNSeventyFive{}) is the full-data ordering, which is why we read that ordering as a property of the representations rather than of the sample size.

\begin{table}[t]
\centering
\caption{Goal-conditioned speed selection. The fitted readout is given a goal distance, selects the push speed it predicts will reach it, and the simulator executes that push with the true physics. \emph{Err.}: median $|$actual $-$ goal$|$; \emph{$\leq$\PlanTol{}\,mm}: fraction of pushes within tolerance. \emph{True $(m,\mu)$}: the same readout on simulator state; \emph{ignore object}: speed alone. Goals are each episode's own true query displacement, so every goal is reachable.}
\label{tab:plan}
\small
\begin{tabular}{lcc cc}
\toprule
& \multicolumn{2}{c}{Clean scene} & \multicolumn{2}{c}{Nuisance variation} \\
\cmidrule(lr){2-3} \cmidrule(lr){4-5}
Representation & Err.\ (mm) & $\leq10$\,mm & Err.\ (mm) & $\leq10$\,mm \\
\midrule
V-JEPA 2 & 2.8 & 94\% & 4.0 & 82\% \\
VideoMAE & 2.8 & 93\% & 7.1 & 64\% \\
VC-1 & 3.2 & 89\% & 13.2 & 40\% \\
DINOv2 & 4.0 & 84\% & 13.5 & 39\% \\
R3M & 2.8 & 93\% & 12.7 & 41\% \\
Random ViT & 4.1 & 82\% & 19.6 & 27\% \\
Frame diff. & 2.7 & 94\% & 5.7 & 71\% \\
Raw pixels & 2.7 & 94\% & 8.5 & 56\% \\
\midrule
True $(m,\mu)$ & 2.9 & 89\% & 2.8 & 91\% \\
Ignore object & 18.0 & 29\% & 19.0 & 28\% \\
\bottomrule
\end{tabular}
\end{table}

\textbf{The gap in task units.} An $R^2$ does not say what a representation is worth to a controller, so we also ask the fitted readout a control-shaped question: given a goal distance, choose the push speed whose predicted displacement is closest, then let the simulator execute that push with the true physics (Table~\ref{tab:main}, right; Table~\ref{tab:plan}). In the clean scene every representation, the random ViT included, lands within a few millimetres of the goal, at the linear readout's own floor (\PlanOracleErr{}\,mm with true $(m,\mu)$). Under nuisance variation V-JEPA~2 misses by \NzPlanVjepaErr{}\,mm and reaches \NzPlanVjepaHit{}\% of goals within 10\,mm, the per-frame encoders miss by roughly $13$\,mm, and the random ViT misses by \NzPlanRandvitErr{}\,mm, matching the object-blind reference (\NzPlanBlindErr{}\,mm). This is open-loop, one step, and simulated, so it is not a policy; its purpose is to express the $R^2$ gap in millimetres.

\begin{table}[t]
\centering
\caption{Encoding versus use. \emph{Probe}: linear decodability of the property from the calibration clips, with frames mean-pooled or, for the per-frame encoders, concatenated. \emph{$\Delta R^2$ after erasing}: change in prediction once the property is made linearly undecodable, in the clean scene and (mass) under nuisance variation. Removing an equal number of random directions costs at most \maxRandomCtl{} (clean) and \NzmaxRandomCtl{} (nuisance).}
\label{tab:probe}
\small
\begin{tabular}{lcccccc}
\toprule
& \multicolumn{2}{c}{Probe $R^2$ (mean-pooled)} & \multicolumn{1}{c}{Concat.} & \multicolumn{2}{c}{$\Delta R^2$ after erasing} & \multicolumn{1}{c}{Nuis.} \\
\cmidrule(lr){2-3} \cmidrule(lr){4-4} \cmidrule(lr){5-6} \cmidrule(lr){7-7}
Representation & Mass & Friction & Friction & Mass & Friction & Mass \\
\midrule
V-JEPA 2 & 0.965 & 0.904 & -- & -0.219 & -0.023 & -0.333 \\
VideoMAE & 0.933 & 0.803 & -- & -0.150 & -0.018 & -0.403 \\
VC-1 & 0.791 & 0.390 & 0.633 & -0.130 & -0.011 & -0.219 \\
DINOv2 & 0.757 & 0.291 & 0.451 & -0.190 & -0.018 & -0.208 \\
R3M & 0.930 & 0.792 & 0.792 & -0.004 & -0.000 & -0.208 \\
Random ViT & 0.762 & 0.361 & 0.672 & -0.025 & -0.003 & +0.000 \\
Frame diff. & 0.986 & 0.963 & -- & -0.306 & -0.030 & -0.373 \\
Raw pixels & 0.983 & 0.940 & -- & -0.001 & -0.000 & -0.350 \\
\bottomrule
\end{tabular}
\end{table}

\textbf{Probes.} Probing is often used to argue that a model knows a property, with any gap between probe accuracy and behaviour attributed to a failure to act on that knowledge. Two interventions show that probe $R^2$ cannot carry this argument in our setting (Table~\ref{tab:probe}). The first is only a preprocessing change: re-encoding the per-frame models with frames concatenated rather than mean-pooled raises the random ViT's friction probe from \RandvitProbeF{} to \SeqRandvitProbeF{}, above VC-1 (\VconeProbeF{}) and DINOv2 (\DinovProbeF{}), while its prediction changes by $\AggRandvitCal{}$. The same frozen network reads as $0.36$ or $0.67$ depending on one line of code that papers rarely report, so a ranking of encoders by probe accuracy is, here, a ranking of aggregation conventions. The second intervention is erasure. Raw pixels and frame differences decode mass to \PixelProbeM{} and \FramediffProbeM{} in the clean scene, yet removing that direction costs frame differences $\FramediffEraseM{}$ and raw pixels $\PixelEraseM{}$. We attribute the difference to redundancy: raw pixels encode the target's position at every sampled frame, so one erased direction leaves the property recoverable from the rest, while frame differences concentrate it. Under nuisance variation that redundancy is gone and the same erasure costs raw pixels $\NzPixelEraseM{}$; R3M moves from $\RthreemEraseM{}$ to $\NzRthreemEraseM{}$. So whether a decoded property is load-bearing depends on what else the scene leaves available, which a probe cannot see. Erasure never returns prediction to the query-only floor, and we read this as the readout depending on a sufficient statistic of the interaction, namely how far the object slid, rather than on an identified $(m, \mu)$; reading a factor out of an outcome is harder than reading the outcome itself.

\textbf{What a benchmark should report.} Three inexpensive checks would have caught every failure described above. The first is an effect-size audit: simulate the perturbation with no model and publish $|\Delta\text{outcome}|$ for every cell. The second is the spread between a random-features baseline, raw pixels, and the best model; if it is a few hundredths, the scene rather than the encoder is setting the score. The third is a swap or an erasure behind any claim that a representation \emph{uses} a property, since decodability is neither necessary nor sufficient for use. None of the evaluations we started from met these.

\section{Limitations}
\label{app:limits}

The environment is deliberately small: rigid boxes, one contact event per clip, a simulator. The numbers characterise this setting. A frozen encoder with a ridge readout measures what is linearly available, which is a lower bound on what fine-tuning could extract, and we evaluate each model's encoder under a shared readout rather than its own rollout machinery, so a model whose native predictor uses physics differently could score differently. The nuisance scene varies viewpoint, lighting, and clutter but not object category or contact type, and its camera has to be wider than the nominal one to keep the workspace in view, so targets also appear smaller; we cannot attribute the drop to any single factor. The erasure intervention removes linearly decodable structure only; a property could survive non-linearly and still be used. Eight representations suffice to show that a scene can conceal a \spreadNz{} spread, not to characterise which pretraining choices produce invariance in general. Finally, the audited perturbation benchmark is our own earlier design; we do not know how common its defects are in published benchmarks, only that they were invisible from its outputs.


\bibliographystyle{plainnat}
\bibliography{references}

@article{vjepa2,
  title={V-JEPA 2: Self-Supervised Video Models Enable Understanding, Prediction and Planning},
  author={Assran, Mahmoud and Bardes, Adrien and Fan, David and Garrido, Quentin and Howes, Russell and Muckley, Matthew and Rizvi, Ammar and Roberts, Claire and Sinha, Koustuv and Zholus, Artem and others},
  journal={arXiv preprint arXiv:2506.09985},
  year={2025}
}

@inproceedings{turpin2024language,
  title={Language Models Don't Always Say What They Think: Unfaithful Explanations in Chain-of-Thought Prompting},
  author={Turpin, Miles and Michael, Julian and Perez, Ethan and Bowman, Samuel R.},
  booktitle={Advances in Neural Information Processing Systems (NeurIPS)},
  year={2023}
}

@article{intphys,
  title={IntPhys: A Framework and Benchmark for Visual Intuitive Physics Reasoning},
  author={Riochet, Ronan and Castro, Mario Ynocente and Bernard, Mathieu and Lerer, Adam and Fergus, Rob and Izard, V{\'e}ronique and Dupoux, Emmanuel},
  journal={arXiv preprint arXiv:1803.07616},
  year={2018}
}

@inproceedings{physion,
  title={Physion: Evaluating Physical Prediction from Vision in Humans and Machines},
  author={Bear, Daniel M. and Wang, Elias and Mrowca, Damian and Binder, Felix J. and Tung, Hsiao-Yu Fish and Pham, R. T. and Haber, Nick and Yamins, Daniel L. K. and others},
  booktitle={NeurIPS Datasets and Benchmarks Track},
  year={2021}
}

@inproceedings{phyre,
  title={PHYRE: A New Benchmark for Physical Reasoning},
  author={Bakhtin, Anton and van der Maaten, Laurens and Johnson, Justin and Gustafson, Laura and Girshick, Ross},
  booktitle={Advances in Neural Information Processing Systems (NeurIPS)},
  year={2019}
}

@inproceedings{vc1,
  title={Where are we in the search for an artificial visual cortex for embodied intelligence?},
  author={Majumdar, Arjun and Yadav, Karmesh and Arnaud, Sergio and Ma, Yecheng Jason and Chen, Claire and Silwal, Sneha and Jain, Aryan and Berges, Vincent-Pierre and Abbeel, Pieter and Malik, Jitendra and Batra, Dhruv and Lin, Yixin and Maksymets, Oleksandr and Rajeswaran, Aravind and Meier, Franziska},
  booktitle={Advances in Neural Information Processing Systems (NeurIPS)},
  year={2023}
}

@inproceedings{r3m,
  title={R3M: A Universal Visual Representation for Robot Manipulation},
  author={Nair, Suraj and Rajeswaran, Aravind and Kumar, Vikash and Finn, Chelsea and Gupta, Abhinav},
  booktitle={Conference on Robot Learning (CoRL)},
  year={2022}
}

@article{dinov2,
  title={DINOv2: Learning Robust Visual Features without Supervision},
  author={Oquab, Maxime and Darcet, Timoth{\'e}e and Moutakanni, Th{\'e}o and Vo, Huy and Szafraniec, Marc and Khalidov, Vasil and Fernandez, Pierre and Haziza, Daniel and Massa, Francisco and El-Nouby, Alaaeldin and others},
  journal={Transactions on Machine Learning Research},
  year={2024}
}

@inproceedings{videomae,
  title={VideoMAE: Masked Autoencoders are Data-Efficient Learners for Self-Supervised Video Pre-Training},
  author={Tong, Zhan and Song, Yibing and Wang, Jue and Wang, Limin},
  booktitle={Advances in Neural Information Processing Systems (NeurIPS)},
  year={2022}
}

@inproceedings{mujoco,
  title={MuJoCo: A physics engine for model-based control},
  author={Todorov, Emanuel and Erez, Tom and Tassa, Yuval},
  booktitle={IEEE/RSJ International Conference on Intelligent Robots and Systems (IROS)},
  year={2012}
}

@inproceedings{galileo,
  title={Galileo: Perceiving Physical Object Properties by Integrating a Physics Engine with Deep Learning},
  author={Wu, Jiajun and Yildirim, Ilker and Lim, Joseph J. and Freeman, William T. and Tenenbaum, Joshua B.},
  booktitle={Advances in Neural Information Processing Systems (NeurIPS)},
  year={2015}
}

@inproceedings{physics101,
  title={Physics 101: Learning Physical Object Properties from Unlabeled Videos},
  author={Wu, Jiajun and Lim, Joseph J. and Zhang, Hongyi and Tenenbaum, Joshua B. and Freeman, William T.},
  booktitle={British Machine Vision Conference (BMVC)},
  year={2016}
}

@inproceedings{hewitt2019control,
  title={Designing and Interpreting Probes with Control Tasks},
  author={Hewitt, John and Liang, Percy},
  booktitle={Empirical Methods in Natural Language Processing (EMNLP)},
  year={2019}
}

@article{belinkov2022probing,
  title={Probing Classifiers: Promises, Shortcomings, and Advances},
  author={Belinkov, Yonatan},
  journal={Computational Linguistics},
  volume={48}, number={1}, pages={207--219},
  year={2022}
}

@inproceedings{agrawal2016poke,
  title={Learning to Poke by Poking: Experiential Learning of Intuitive Physics},
  author={Agrawal, Pulkit and Nair, Ashvin and Abbeel, Pieter and Malik, Jitendra and Levine, Sergey},
  booktitle={Advances in Neural Information Processing Systems (NeurIPS)},
  year={2016}
}

@inproceedings{kumar2021rma,
  title={RMA: Rapid Motor Adaptation for Legged Robots},
  author={Kumar, Ashish and Fu, Zipeng and Pathak, Deepak and Malik, Jitendra},
  booktitle={Robotics: Science and Systems (RSS)},
  year={2021}
}

@article{elazar2021amnesic,
  title={Amnesic Probing: Behavioral Explanation with Amnesic Counterfactuals},
  author={Elazar, Yanai and Ravfogel, Shauli and Jacovi, Alon and Goldberg, Yoav},
  journal={Transactions of the Association for Computational Linguistics},
  volume={9}, pages={160--175},
  year={2021}
}

@inproceedings{ravfogel2020null,
  title={Null It Out: Guarding Protected Attributes by Iterative Nullspace Projection},
  author={Ravfogel, Shauli and Elazar, Yanai and Gonen, Hila and Twiton, Michael and Goldberg, Yoav},
  booktitle={Annual Meeting of the Association for Computational Linguistics (ACL)},
  year={2020}
}

\appendix
\section{Effect-size audit}
\label{sec:audit}

The single-perturbation recipe correlates representational distance with outcome change, which presupposes that the perturbation changes the outcome. We simulate matched pairs across objects and seeds for four tasks and five properties and record $|\Delta\text{outcome}|$ with no model involved (Table~\ref{tab:audit}).

\auditDead{} of \auditTotal{} cells are inert, for three reasons. In two of four tasks the object is released above a surface and falls; the recorded outcome is its distance to a goal, a geometric constant. Neither mass nor friction affects a free fall. The rigidity axis set \texttt{solref="-1000 1"}, which mixes MuJoCo's two parameterisations; the simulator discarded the value and used its default, so every rigid-versus-deformable pair was the same scene, and the run completed with one warning per scene. Stacking was scored by the upper object's final height, which geometry fixes once the stack settles, so the dynamics a perturbation does change (bounce, settling, toppling) were not measured.

\begin{table}[h]
\centering
\caption{Effect-size audit of a single-perturbation benchmark: mean $|\Delta\text{outcome}|$ as a percentage of the outcome scale. \dead{} marks cells where the perturbation moves the outcome by less than $10^{-4}$, within integrator noise.}
\label{tab:audit}
\small
\begin{tabular}{lccccc}
\toprule
Task & Mass & Friction & Rigidity & Size & Height \\
\midrule
pick\_and\_place & \dead & \dead & \dead & \dead & \dead \\
push\_to\_goal & 149\% & 70\% & \dead & 16\% & \dead \\
stack\_two & $<$1\% & $<$1\% & \dead & 22\% & \dead \\
insert\_peg & \dead & \dead & \dead & 33\% & \dead \\
\bottomrule
\end{tabular}
\end{table}

None of these is visible from a results table. In the same codebase, two encoder wrappers substituted randomly initialised networks when a checkpoint failed to load, and one of those checkpoints did not exist, so a column reported as a physics-aware world model was random features. Our encoders raise on load failure.

\section{Environment design details}
\label{app:design}

\textbf{Collisions rather than pushes.} A force $F$ applied to mass $m$ against sliding friction $\mu$ moves the object only if $F/m > \mu g$, and displacement then grows as $(F/m - \mu g)^2$. Over a mass range wide enough to matter, a fixed force leaves roughly half the objects stationary and sends the rest out of frame; sampling forces and masses independently and requiring all three phases to land in a $[1, 30]$\,cm window accepted under $2\%$ of draws. A collision has no threshold, recoil speed is bounded by impact speed, and the same window accepts nearly every draw.

\textbf{No rolling shapes.} A sphere carries its recoil on rolling contact, so surface friction barely decelerates it. At matched physics spheres travelled $5.6\times$ farther than boxes (median $0.56$ versus $0.06$\,m) and $\mathrm{corr}(\mu, \text{displacement})$ was $+0.10$ for spheres against $-0.39$ for boxes, so friction would not be identifiable from the outcome. Targets are boxes of varying aspect ratio.

\textbf{Timing.} Impact speeds and friction are chosen jointly so a slide lasts about $0.4$\,s rather than $0.15$\,s at the same displacement, spreading the motion over about 13 rendered frames instead of 5. Episodes whose displacement falls below one rendered pixel ($<12$\,mm) or leaves the workspace ($>28$\,cm) are resampled; acceptance exceeds $70\%$.

\textbf{Nuisance variation.} Camera pose, lighting, floor tint, and two distractor objects are resampled per clip. Every draw is checked to keep the workspace in view, since a camera that crops the slide would make the episode unanswerable and register as a representation failure. Meeting that constraint requires a wider field of view than the nominal camera (scale $1.14$--$1.30$).

\textbf{Readout details.} Each clip's features form one block. Blocks are centred and reduced by PCA to a shared component budget on the training rows, standardised, concatenated with the commanded speed, and fit by ridge regression with the penalty selected by inner cross-validation. Repeated 5-fold cross-validation (3 repeats) gives the reported means and fold spreads. Probes use the same pipeline with the latent property as target. Erasure follows iterative nullspace projection: probe directions are removed until a tuned probe's $R^2$ for the property is at chance, and prediction is re-fit on the projected features; the control removes the same number of random directions. Speed selection uses the calibration-only readout (commanded speed plus the two calibration clips), since the query clip is rendered at the true query speed and cannot be re-rendered for each candidate; the speed is chosen on a 451-point grid over $[0.40, 0.85]$\,m/s.

\textbf{Seed robustness.} Repeating the evaluation with a different cross-validation seed moves every reported quantity by at most $0.01$; for DINOv2 in the clean scene, prediction with calibration $0.983 \to 0.983$, friction probe $0.345 \to 0.346$, and the mass-erasure effect $-0.227 \to -0.232$.

\end{document}